\documentclass[sigconf,screen]{acmart}

\copyrightyear{2021}
\acmYear{2021}
\setcopyright{acmlicensed}\acmConference[SSTD '21]{17th International Symposium on Spatial and Temporal Databases}{August 23--25, 2021}{virtual, USA}
\acmBooktitle{17th International Symposium on Spatial and Temporal Databases (SSTD '21), August 23--25, 2021, virtual, USA}
\acmPrice{15.00}
\acmDOI{10.1145/3469830.3470906}
\acmISBN{978-1-4503-8425-4/21/08}

\usepackage{amsmath,amsfonts}

\usepackage{amssymb}
\usepackage{algorithmic}
\usepackage{graphicx}
\usepackage{textcomp}
\usepackage{xcolor}
\usepackage{url}
\usepackage{subcaption}
\usepackage{enumitem}
\usepackage{booktabs}
\usepackage{textcomp}
\usepackage{xspace}
\usepackage{soul}

\graphicspath{{figs/}}

\newcommand{\srec}{S$^3$Rec\xspace}
\newcommand{\ours}{M$^3$Rec\xspace}
\newcommand{\lego}{LEGO\textregistered Cube\xspace}

\newcommand{\mysec}[1]{Section~#1}
\newcommand{\myfig}[1]{Figure~#1}

\newcommand{\mytabs}[1]{Tables~#1}
\newcommand{\mydef}[1]{Definition~#1}
\newcommand{\myeq}[1]{Equation~#1}

\begin{document}

\title{Sequential Recommendation in Online Games with Multiple Sequences, Tasks and User Levels}

\author{Si Chen}
\email{sichen@stu.xmu.edu.cn}
\authornote{Work done when the first author was an intern in Tencent.}
\affiliation{%
  \institution{School of Informatics, Xiamen University}
  \city{Xiamen}
  \country{China}
}

\author{Yuqiu Qian}
\email{yuqiuqian@tencent.com}
\affiliation{%
  \institution{Tencent}
  \city{Shenzhen}
  \country{China}
}

\author{Hui Li}
\email{hui@xmu.edu.cn}
\authornote{Hui Li and Chen Lin are the corresponding authors. Email: hui@xmu.edu.cn, chenlin@xmu.edu.cn.}
\affiliation{%
  \institution{School of Informatics, Xiamen University}
  \city{Xiamen}
  \country{China}
}

\author{Chen Lin}
\email{chenlin@xmu.edu.cn}
\authornotemark[2]
\affiliation{%
  \institution{School of Informatics, Xiamen University}
  \city{Xiamen}
  \country{China}
}

\renewcommand{\shortauthors}{S. Chen et al.}

\begin{abstract}
Online gaming is growing faster than ever before, with increasing challenges of providing better user experience. Recommender systems (RS) for online games face unique challenges since they must fulfill players' distinct desires, at different user levels, based on their action sequences of various action types. Although many sequential RS already exist, they are mainly single-sequence, single-task, and single-user-level. In this paper, we introduce a new sequential recommendation model for multiple sequences, multiple tasks, and multiple user levels (abbreviated as M$^3$Rec) in Tencent Games platform, which can fully utilize complex data in online games. We leverage Graph Neural Network and multi-task learning to design M$^3$Rec in order to model the complex information in the heterogeneous sequential recommendation scenario of Tencent Games. We verify the effectiveness of M$^3$Rec on three online games of Tencent Games platform, in both offline and online evaluations. The results show that M$^3$Rec successfully addresses the challenges of recommendation in online games, and it generates superior recommendations compared with state-of-the-art sequential recommendation approaches.
\end{abstract}

\ccsdesc[500]{Information systems~Recommender systems}

\keywords{online games, sequential recommender systems, multi-task learning, graph neural network}

\maketitle


\section{Introduction}
\label{sec:intro} 

As web services are ever-expanding, sequential data (e.g., users' click logs
or user traveling history) become prevalent in Recommender Systems (RS) and therefore
\emph{Sequential Recommender Systems} (SRS) have attracted more and more
attention~\cite{LudewigJ18,QuadranaCJ18,abs-1905-01997}. Given users'
historical behavior sequences, SRS aim at predicting his/her next action,
e.g., the next point of interest (i.e., POI) to visit, or the next product
he/she will buy. 
Unlike traditional RS which learn from the \emph{two-way}
user-item interactions, SRS model the \emph{three-way} interaction among a
user, an item he/she has selected and an item he/she will select next. 
One
specific application of SRS is recommendation in online games which is growing faster than ever before, with increasing challenges of providing better user experience. In this paper, we study the sequential recommendation problem in \emph{Tencent
Games}\footnote{https://game.qq.com} where the heterogeneous
information makes it difficult for traditional SRS to fully model the sequential behaviors of users.

\begin{figure}[!t]
\centering
\begin{subfigure}{1\columnwidth}
  \centering
  \includegraphics[width=0.95\columnwidth]{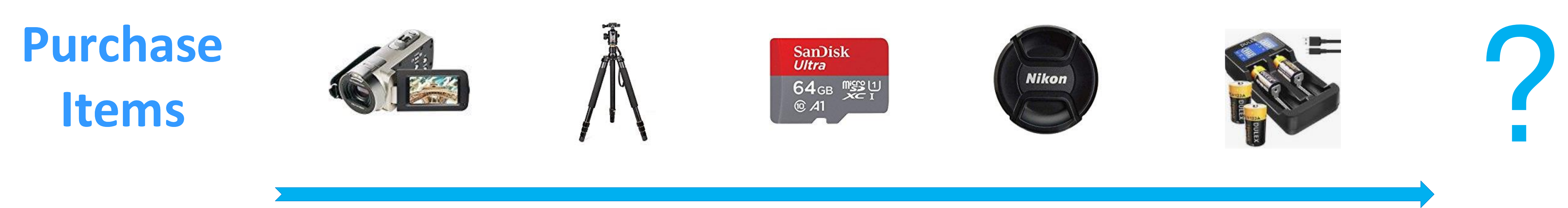}
  \caption{Traditional \srec.}
  \label{fig:single_seq}
\end{subfigure}%
\vspace{10pt}
\begin{subfigure}{1\columnwidth}
  \centering
  \includegraphics[width=0.95\columnwidth]{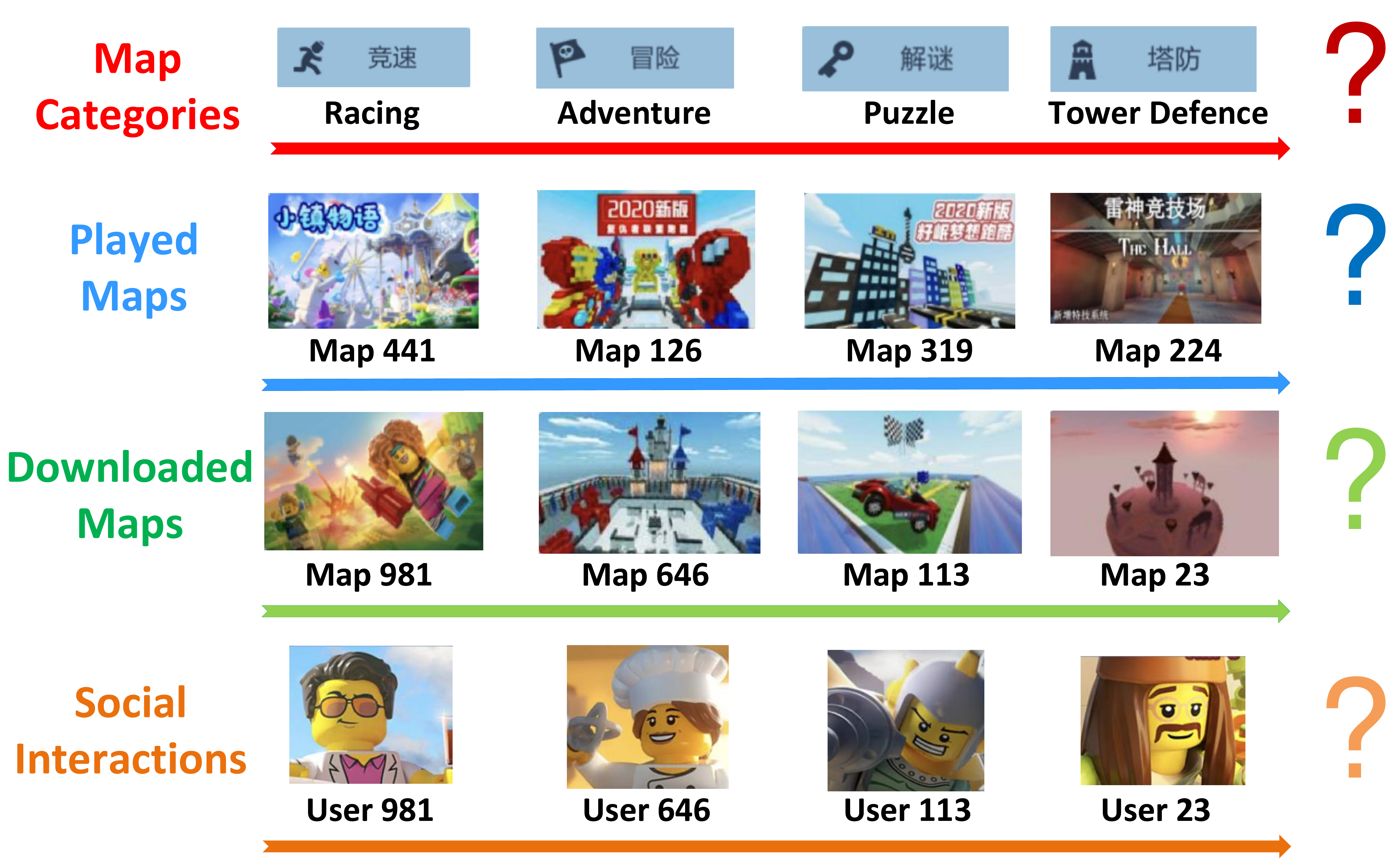}
  \caption{\ours for \lego.}
  \label{fig:multi_seq}
\end{subfigure}
\caption{A comparison between traditional \srec and \ours for Tencent online game \lego.}
\label{fig:comp_seq}
\end{figure}

Although many SRS already exist~\cite{LudewigJ18,QuadranaCJ18,abs-1905-01997}, they are
mainly single-sequence, single-task, and single-user-level (i.e., modeling
individual user instead of user groups), i.e., \emph{single-task learning at
single-user level for single-sequence RS} (\srec for short). 
\myfig{\ref{fig:single_seq}}
demonstrates an example of such SRS. The blue timeline indicates the purchase sequence, i.e., products that one user bought. The task is to recommend the next item to purchase for the user. As the previous items are related to video cameras, the next item is likely to be also related to video camera. 
Some recent
methods consider more content features (e.g., temporal feature, text
and item category)~\cite{HidasiQKT16,TuanP17} to enhance SRS's ability. Though heterogeneous information has been considered, these methods
still model sequential behaviors based on a single sequence. Additional
information is utilized as auxiliary data and incorporated into the
single sequence to optimize for a single task, i.e., predicting the next item the user will interact with. 

\srec is suitable for recommendation scenarios such as book or movie recommendation. However, the recommendation in online games faces unique challenges that \srec cannot handle well:
\begin{itemize}
\item Firstly, the action types are diverse in online gaming platforms. In other recommendation scenarios, such as book or movie recommendation, users can only adopt limited actions, e.g., purchase or click. However, in online games, players are able to adopt many types of actions. For
example, \myfig{\ref{fig:multi_seq}} depicts part of the recommendation scenario
of the game \lego\footnote{https://lgwx.qq.com} in Tencent
Games platform, where
heterogeneous information (e.g, map category, played maps, downloaded maps,
and social interactions) is contained in \emph{multiple user action sequences} and they have different meanings for the same user. Conventional SRS are unable to distinguish such differences.

\item Secondly, there are multiple tasks to be fulfilled in online gaming platforms. In other recommendation scenarios, recommendation providers are only interested in the main action type, e.g., purchase. However, predicting the next item to purchase is not the only task in online games. For example,  in \lego, the \emph{multiple tasks} of predicting next map to download, next map to play are all important for the game developer and publisher to efficiently allocate resource, improve user experience, and increase revenue.

\item Finally, there is different representation levels for users and user groups in online gaming platforms. Most online games incorporate strong social factors that allow users to team up in battles and regions. Thus, in order to deliver more accurate recommendations, SRS for online games need to model not only single-level user representations (i.e., representations for each single user) but also
\emph{multiple user-level} representations (i.e., representations for user groups). However, few existing works consider modeling user representations in sequential
recommendation, as pointed out by Fang et al.~\cite{abs-1905-01997}. 
\end{itemize}

The new challenges in SRS for Tencent Games and the limitations of previous
SRS motivate us to propose a \emph{sequential recommender system with
multiple sequences, tasks and user levels} (abbreviated as \ours). We
leverage Graph Neural Network (GNN)~\cite{WuPCLZY20} and multi-task learning~\cite{ZhangY17aa} to
design a multi-sequence, multi-task and multi-level neural recommendation
architecture for \ours to fully utilize the complex information in the
heterogeneous sequential recommendation scenario of Tencent Games. More descriptions 
for the recommenders in Tencent Games are provided in \mysec{\ref{sec:pre}}.

In summary, the major contributions of this paper are:
\begin{itemize}
	\item We identify and formally define a new multi-sequence, multi-task and multi-level sequential recommendation problem, which is a practical problem existing in online games of Tencent Games platform. To our best knowledge, we are first to study this problem in terms of sequential recommendation in online games.
	\item We propose a new framework \ours for sequential learning in the heterogeneous recommendation scenario. \ours has the capacity to model the complex meanings of multiple sequences beneath the surface.
	\item We adopt multi-task learning in \ours so that it is able to optimize several tasks in parallel.
	\item We evaluate \ours on real data of three online games in Tencent Games platform and verify the effectiveness of \ours for the traditional task of recommending next item in SRS. Meanwhile, \ours can offer suggestions for other prediction tasks in online games with high quality.
\end{itemize}

The rest of the paper is organized as follows: \mysec{\ref{sec:pre}} gives an
overview of the recommendation scenario in Tencent Games and formally
defines the new recommendation problem. We illustrate our proposed method, \ours, in
\mysec{\ref{sec:model}}. In \mysec{\ref{sec:exp}}, we compare \ours with other
state-of-the-art sequential recommendation algorithms and verify its
effectiveness. \mysec{\ref{sec:con}} concludes our work.

\section{Sequential Recommender System for \lego}
\label{sec:pre}

In this section, we will give an overview of \lego as an example to illustrate the recommendation scenario of online games in Tencent Games platform. Then, we will define the sequential recommendation problem for \ours in online games.

\subsection{Overview of SRS for \lego}
\label{sec:sigma} 

\lego is a popular sandbox game developed and operated by Tencent Games. It provides a virtual world comprised of different areas for Chinese mobile players. Players are encouraged to freely download area maps, explore these areas, construct with LEGO bricks, and get involved in various tasks with friends. As such, for each player, four behavior sequences are available: the \emph{map sequence} containing game maps that each user has downloaded, the \emph{type category} containing map categories of each user's downloaded maps, the \emph{play sequence} containing game maps played by each player, and the \emph{friend sequence} containing other users that each user has interacted with. 

\myfig{~\ref{fig:multi_seq}} illustrates how the SRS in \lego organizes the
data of multiple sequences for one player:
\begin{itemize}
  \item The green timeline contains the downloaded map sequence of this player. It corresponds to the main task, namely recommending the next map that the target player will download.
  \item The red timeline contains the map categories corresponding to maps in the green timeline. The task for this sequence is to predict the map category of the next map to be recommended. This task is helpful for game operation engineers when the information of some players is insufficient to make fine-grained recommendations and they can use the category recommendation instead of map recommendation.
  \item The blue timeline contains the played map sequence of this player. The auxiliary task for this sequence is to predict which map will be played by the player next. As maps must be firstly downloaded and then they can be played, this task is useful for distinguishing maps which players will play even without being recommended in the download sequence (i.e., main task) and maps that players will play after they get exposure via the recommendation. Game operation engineers may tune the recommendation priority for the latter in the main task so that they can be played. 
  \item The orange timeline contains the user-user interaction sequence of this player. The auxiliary task for this sequence is to predict the next player that this player will interact with. The predication results of this sequence can help game operation engineers identify similar or relevant players, which is used as part of the input to the game player clustering in \mysec{\ref{sec:group}}.
\end{itemize}

The meanings
of the elements in each sequence are different under the surface. 
For example, 
one map
appearing in green timeline indicates that the player has downloaded the map, but
it is not the indication of his/her preferences. Since SRS in the game may push maps
to players based on some strategies (e.g., advertising), it is possible that
players will download some maps that he/she does not feel interesting actually. As a comparison, when a player played a map 
 in blue timeline, the system can understand that it is a strong and
explicit indication of his/her preferences. 
In summary, there are five recommendation problems which naturally arise from the sequences shown in \myfig{\ref{fig:multi_seq}}:
\begin{enumerate}
\item \textbf{Next map prediction} is the main task that predicts which map will be downloaded by the user next.
\item \textbf{Next category prediction} is an auxiliary task that predicts which category of map will be downloaded by the user next.
\item \textbf{Next play prediction} is an auxiliary task that predicts which map will be played by the user next.
\item \textbf{Next friend prediction} is an auxiliary task that predicts who will be added to the user's friend list next.
\item \textbf{Next group map prediction} is an auxiliary task that predicts the next download map for a user group. 
\end{enumerate}
The last task is related to a concept of \emph{multiple user levels} in online games. Game operation engineers typically group users by group characters and sometimes promote game activities to some specific user group. In this case, the recommendation is perform on the second user level, i.e., user group, instead of the first user level, i.e., individual player.

Different sequences in \lego involve different meanings of
players' behaviors and simply incorporating them into one sequence without
distinction will result in information loss. On the other hand, prediction
tasks in \lego vary in their goals and optimizing them
simultaneously in one sequence and single user level is not straightforward.
Therefore, previous sequential recommendation algorithms, which study single
sequence, task and user level, cannot model the complex scenario in \lego.

\subsection{Problem Definition}

We now first give the formal definition of the recommendation task of \srec:

\begin{definition}[Recommendation Task of \srec]
\label{def}
Let $I=\{i_1,...\,, i_{|I|}\}$ be a set of items. The action sequence
$s$ for a user is the sequence of items that the user has
interacted with: 
$\{a_1,...\,,a_{|s|}\}$, where $a_j=\langle i_j, t_j\rangle$
indicates that the user selected item $i_j$ at time $t_j$ with
$t_{|s|}\geq t_{|s|-1}\geq...\geq\,t_{1}$, and $|s|$ is the number of actions in $s$. Given an
action sequence $s$, \srec predicts the item(s)
that the user will next add to $s$. Items can be
replaced by users or item categories if the sequence is related to user-user interactions or item information like categories.
\end{definition}

We can extend \mydef{\ref{def}} and
define the recommendation task of \ours:

\begin{definition}[Recommendation Task of \ours]
\label{def_2} 
Let $I=\{i_1,...\,, i_{|I|}\}$ be a set of items. The action sequence set
$\mathcal{S}=\{s_1,\cdots,s_m\}$ for a user (or a user group)
has $m$ sequences (one main sequence and $m-1$ auxiliary sequences) with possible different lengths. Each sequence
$s_q$ consists of items that the user (or user group) has interacted with: $\{a_1,...\,,a_{|s_q|}\}$, where $a_j=\langle
i_j, o, t_j\rangle$ indicates that the user (or user group) performed action $o$ on item $i_j$ at time $t_j$, and $t_{|s_q|}\geq
t_{|s_j|-1}\geq...\geq\,t_{1}$. Given
$\mathcal{S}$, the main task of \ours is to predict the item in the next action that the user (or user group) will next add to the main
sequence. Other auxiliary tasks aim at predicting the item/item category/user in the next action that the user (or user group) will next add to the
auxiliary sequences. 
\end{definition}

Similar to \srec, \ours does not predict the timestamp for the next
action. Besides, the action type $o$ is the same for all the actions in one
sequence. Therefore, \ours does not need to predict the action
type for new action. 


\section{\ours for Tencent Games Platform}
\label{sec:model}
In this section, we will first review the Gated Graph Sequence Neural Network~\cite{LiTBZ15} based \srec~\cite{WuT0WXT19} since \ours is built on the top of it. Then we will describe how \ours is able to model the sequential recommendation problem with multiple sequences, tasks and user level which is the scenario that online games in Tencent Games Platform have. \myfig{\ref{fig:model}} provides an overview of \ours.

\subsection{GGS-NN Based \srec}
\label{sec:single_rec}

\begin{figure*}[!t]
\centering
\includegraphics[width=1\textwidth]{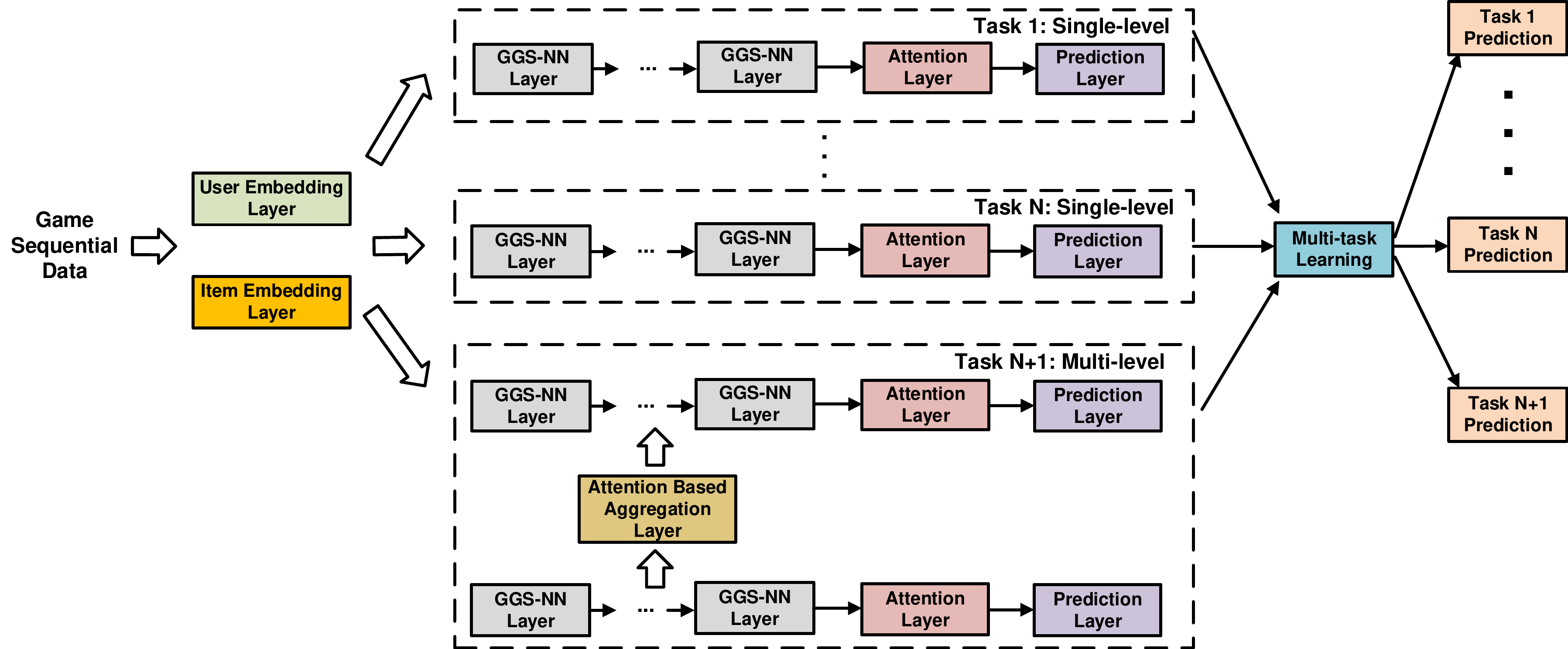}
\caption{Overview of \ours for online games.}
\label{fig:model}
\end{figure*}

Li et al.~\cite{LiTBZ15} introduce the gate mechanism, which has been shown to
be effective in Long Short-term Memory, into Graph Neural Network (GNN)~\cite{ScarselliGTHM09} and propose the Gated Graph Sequence
Neural Network (GGS-NN) for modeling graphs as sequences. Later, Wu et
al.~\cite{WuT0WXT19} apply GGS-NN to \srec. In \ours, we use their design for the single-sequence, single-task and single-user-level recommendation unit in \ours, i.e., each of the tasks $1$ to $N$ in \myfig{\ref{fig:model}}.

\begin{figure}[!t]
\centering
\includegraphics[width=1\columnwidth]{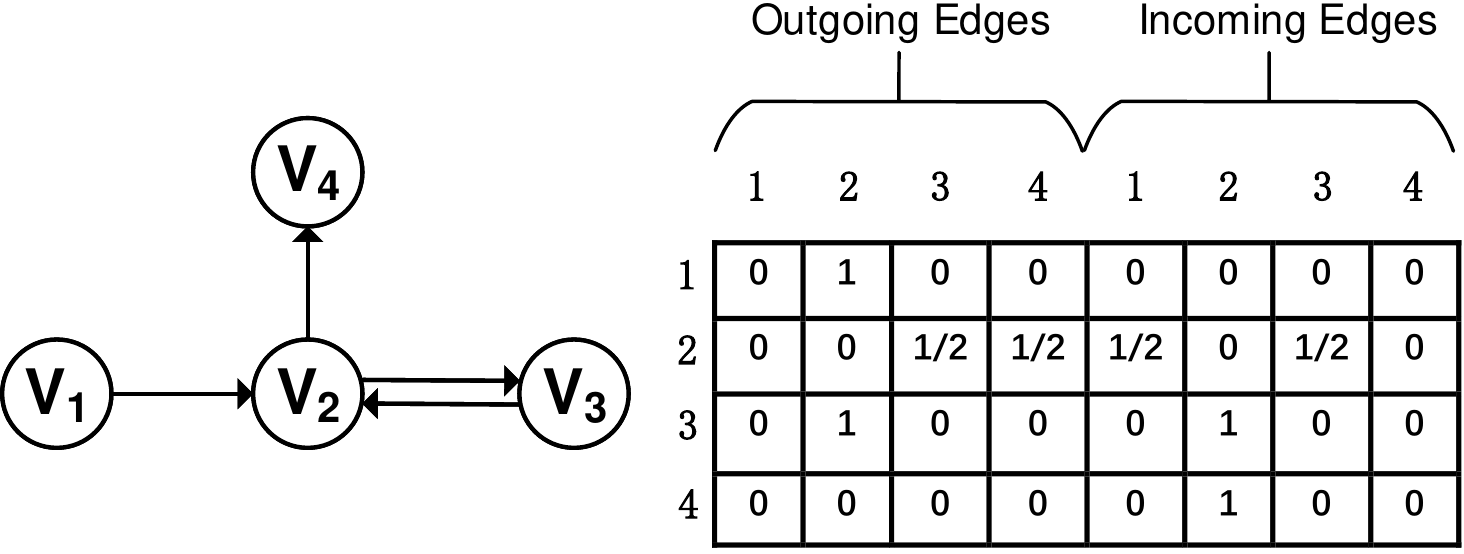}
\caption{An example of a sequence graph and the connection matrix $\mathbf{A}$.}
\label{fig:sequence_graph}
\end{figure}

Specifically, each user action sequence $s$ can be viewed as a sequence graph $\mathcal{G}_s$ and the update rule for the representation $h_i$ of each node (i.e., item) $i$ in $s$ from epoch $t-1$ to epoch $t$ using a GGS-NN layer can be defined as follows:
\begin{equation}
\begin{aligned}
\mathbf{c}_{s,i}^{(t)}&=\mathbf{A}_{s,i:}^T\left[ \mathbf{h}_1^{(t-1)},\cdots,\mathbf{h}_n^{(t-1)} \right]^T + \mathbf{b}\\
\mathbf{z}_{s,i}^{(t)}&=\sigma(\mathbf{W}_{z}\mathbf{c}_{s,i}^{(t)}+\mathbf{V}_{z}\mathbf{h}_{i}^{(t-1)})\\
\mathbf{r}_{s,i}^{(t)}&=\sigma(\mathbf{W}_r\mathbf{c}^{(t)}_{s,i}+\mathbf{V}_{r}\mathbf{h}_i^{(t-1)})\\
\tilde{\mathbf{h}}^{(t)}_{i}&=tanh \big(\mathbf{W}_h\mathbf{c}_{s,i}^{(t)}+\mathbf{V}_{h}(\mathbf{r}_{s,i}^{(t)}\odot\mathbf{h}_{i}^{(t-1)}) \big)\\
\mathbf{h}_{i}^{(t)}&=(1-\mathbf{z}_{s,i}^{(t)})\odot \mathbf{h}_{i}^{(t-1)}+\mathbf{z}_{s,i}^{(t)}\odot \tilde{\mathbf{h}}_{i}^{(t)}
\end{aligned}
\end{equation}
where $\mathbf{c}$, $\mathbf{z}$, $\mathbf{r}$ and $\tilde{\mathbf{h}}$ are gathering gate, reset gate, update gate, and candidate activation, respectively. $\sigma(\cdot)$ is the sigmoid function, $\mathbf{h}_{*}$ is the vector representation of node in sequence $s$, ``$\odot$'' is the element-wise multiplication, and $\mathbf{W}_{*}$ and $\mathbf{V}_{*}$ are learnable matrices. $\mathbf{A}_s\in \mathbb{R}^{n_s\times 2n_s}$ is the connection matrix for $s$ (i.e., the concatenation of outgoing adjacent matrix and ingoing adjacent matrix), and $n_s$ indicates the number of unique nodes in $s$. $\mathbf{A}_{s,i:}\in \mathbb{R}^{n_s\times 2}$ are the two columns of blocks in $\mathbf{A}_s$ with respect to node $i$. For instance, \myfig{\ref{fig:sequence_graph}} illustrates an example showing the connection matrix for one user's user-item interaction sequence $\{ v_1, v_2, v_3, v_2, v_4\}$. 

We may stack several GGS-NN layers to enhance the non-linearities of the model and we call it \emph{GGS-NN unit} in the following. For one user's user-item interaction sequence $\{ v_1, \cdots, v_l\}$ with $l$ items in the sequence, the prediction for the next item is generated based on the local sequence embedding (i.e., the learned representation $\mathbf{h}_l$ for the last node $v_l$ in the sequence) and the global sequence embedding $\mathbf{s}_g$ which is captured by a soft-attention layer:
\begin{equation}
\alpha_i=\mathbf{g}^T\sigma(\mathbf{W}_1\mathbf{h}_l+\mathbf{W}_2\mathbf{h}_i+\mathbf{c}),\,\,\,\mathbf{s}_g = \sum_{i=1}^{l}\alpha_i\mathbf{h}_i
\end{equation}
where $\mathbf{g}$, $\mathbf{W}_1$ and $\mathbf{W}_2$ are learnable weights. $\mathbf{b}$ is the bias vector. 

Finally, the predicted probability of an item $j$ being the next item is:
\begin{equation}
\label{eq:predict}
\hat{r}_j=\big(\mathbf{W}_3 (\mathbf{s}_g \oplus \mathbf{h}_l)\big)^T\mathbf{v}_j,
\end{equation}
where $\mathbf{v}_j$ is the embedding for the item $j$, $\mathbf{W}_3$ is a learnable weight matrix, and ``$\oplus$'' indicates the concatenation operation.

\subsection{Modeling Multiple Sequences and Multi Tasks in \ours}

Considering the multi-sequence and multi-task scenario that online games face, we choose to adopt multiple GGS-NN units for online games in Tencent Game platform. Each of them is used for a specific sequence and a specific task. Items in the original GGS-NN unit are replaced with item categories or users according to the task. For next category prediction task, a separate map category embedding layer is used instead of the item embedding layer. 

As shown in \myfig{\ref{fig:sequence_graph}}, each task is modeled using an independent GGS-NN unit with the soft-attention mechanism, and all tasks share the user embedding layer and item embedding layer. For instance, the same user $i$ in the next map prediction task and the next friend prediction task of \lego will share the same user embedding $\mathbf{u}_i$  encoded by the shared user embedding layer. All the GGS-NN unit are trained together in the manner of multi-task learning. The details of the optimization will be illustrated in \mysec{\ref{sec:opt}}.

\subsection{Modeling Multiple User Levels in \ours}
\label{sec:group}

Game operation engineers lead game players to particular game activities for a better user experience, and a game activity typically has a target user group. Promoting game activities may increase the chance for users to see the maps they favor. 
Game players in \lego have group characters, and operations engineers regularly update the user statistics. We run k-means clustering algorithm over these data to construct the player grouping information.

After identifying the player groups in the game, we are actually dealing with
a classical problem in the recommender system community, namely \emph{group
recommendation}~\cite{Amer-YahiaRCDY09,RoyLL15}, in the sequential
recommendation task in Tencent Games platform. In traditional group
recommendation task, the system aims at recommending items to a group of users
that maximize the satisfaction of the group members, according to some
semantics of group satisfaction. Users in the same group share some
similarities (e.g., tastes or locations), and all members in the same group
receive the same recommendation from the system. Group recommendation has been
widely studied and applied in social recommenders (e.g.,
Shelfari~\cite{LiWTM15}), event recommenders (e.g.,
Plancast~\cite{LuLMC17}), restaurant recommenders (e.g.,
ZAGAT~\cite{NtoutsiSNK12}), just to name a few. However, only a few studies~\cite{PiliponyteRK13,StratigiNPS20} has considered the group recommendation problem in sequential recommender. 

We first define the \emph{Hierarchical Multi-level Recommendation} for online games. Formally, we have a two-level recommendation architecture as shown in Task $N+1$ of \myfig{\ref{fig:model}}. On the bottom level, \ours conducts the general single-level sequential recommendation task, i.e., recommend an item to each individual game player. On the second level, \ours turns to \emph{sequential group recommendation} task, i.e., recommend an item to groups of game players.

We design an attention based aggregation layer to aggregate the information from each group member at the bottom level and then construct the representation of the group on the second level.
Recall that $\mathbf{h}_i$ indicates the representation of item $i$. Suppose that $\mathbf{q}_j$ is the representation of player $j$ on the first level, $\mathbf{p}_g$ is the representation of player group $g$ at the bottom level, and $\mathcal{N}(g)$ indicates all the players that player group $g$ contains. We aggregate the representations of individual players and generate the embedding for group $g$ on the second level: 
\begin{equation}
\label{eq:ho_group}
\begin{aligned}
\mathbf{p}_g &= \text{RELU}\big(\mathbf{W}_p\sum_{i\in \mathbf{M}_g} \beta_{i} \mathbf{h}_i \big)\\
\beta_i &= \frac{exp(\hat{\beta}_i)}{\sum_{i'\in \mathcal{N}(g)}exp(\hat{\beta}_{i'})}\\
\hat{\beta}_i &= \mathbf{W}_{b} \mathbf{e}_{i} + b_{e}\\
\mathbf{e}_i &= \text{RELU}\big(\mathbf{W}_{e} \mathbf{h}_{i} + \mathbf{b}_{e}\big)
\end{aligned}
\end{equation}
where $RELU(\cdot)$ is the Rectified Linear Unit, $\mathbf{W}_{*}$ is a weight matrix, $\mathbf{b}_{e}$ is a bias vector, and $b_{e}$ is a bias term. In \myeq{\ref{eq:ho_group}}, we assign different attention weight $\beta$ to indicate the differing importance of each group member to the group. The attention weight $\beta$ is parameterized with a single-layer feedforward neural network and then normalized by the softmax function. The motivation is that differences still exist among game players belonging to the same group, though players in the same group share very similar characteristics. Considering such nuances helps model the profile of the whole group better.

After obtaining the group representations $\mathbf{p}$, the group recommendation process at the second level is similar to the process of GGS-NN based \srec illustrated in \mysec{\ref{sec:single_rec}}.

\subsection{Optimization of Multi-task Learning}
\label{sec:opt}

For each individual task, the predicted probability distribution for each possible item/item category/user being the next in an action sequence is similar to \myeq{~\ref{eq:predict}}:
\begin{equation}
\label{eq:softmax}
\mathbf{\hat{r}}=\big(\mathbf{W}_3(\mathbf{s}_g\oplus\mathbf{h}_l)\big)^T\mathbf{Q},
\end{equation}
where $\mathbf{Q}$ is all the item embeddings, all the item category embeddings, or all the user embeddings, depending on the detailed task.

Then, the cross-entropy loss is used in the optimization for each task $i$:
\begin{equation}
\label{eq:loss}
\mathcal{L}_i = -\sum_{s\in \mathcal{S}_i}\mathbf{r}_{s} \log(\mathbf{\hat{r}}_{s}),
\end{equation} 
where $\mathcal{S}_i$ is the interaction sequence set for task $i$, $\mathbf{r}_{s}$ indicates the one-hot encoding vector of the ground-truth next item/item category/user in the sequence $s$, and $\mathbf{\hat{r}}_s$ is calculated using \myeq{\ref{eq:softmax}}.

We use a common method to perform multi-task learning in \ours, i.e., assign a task weight to each task loss and optimize the joint loss:
\begin{equation}
\label{eq:joint_loss}
\mathcal{L} = \sum_{i=1}^{t} w_i \mathcal{L}_i,
\end{equation}
where $w_i$ is the task weight for task $i$.
Stochastic gradient descent based methods can be used for the optimization and we use Adam~\cite{KingmaB14} for \ours.


\section{Experiment}
\label{sec:exp}

In this section, we move forward to evaluate the effectiveness of \ours. We aim to answer the following questions:
\begin{itemize}[leftmargin=20pt,topsep=2pt]
	\item[\textbf{RQ1}] How does each task contribute to \ours's performance?
	\item[\textbf{RQ2}] How does \ours perform compared to the state-of-the-art methods in offline datasets?
	\item[\textbf{RQ3}] How does \ours perform in online recommendation test?
\end{itemize}

\subsection{Experimental Setup}

We evaluate the performance of \ours and other competitors in three online games on Tencent Games platform. The first game is \lego and we have illustrated the details of each sequences and each tasks in \mysec{\ref{sec:pre}}. We also select two popular Tencent games belonging to the Role Playing Game (RPG) genre. Both games allow players to team up and combat with each other. Players can evolve their skills by purchasing items. We are required by our industry partners to anonymize the two games as Tgame and Ygame. For each player in Tgame, three behavior sequences are available: the \emph{item sequence} each user has downloaded, the \emph{type sequence} of each user's downloaded item,  and the \emph{friend sequence} each user has interacted with. It presents three recommendation problems: i.e., \emph{next item prediction}, \emph{next type prediction} and \emph{next friend prediction}.
For each player in Ygame, five behavior sequences are available: the \emph{item sequence}, the \emph{type sequence},  the \emph{friend sequence}, the \emph{evolve sequence} of items each user has utilized to evolve his/her skills, and the \emph{buy sequence} of items each user has bought. It presents four recommendation problems: i.e., \emph{next item prediction}, \emph{next type prediction} and \emph{next friend prediction}.
It present five recommendation problems: i.e., \emph{next item prediction}, \emph{next type prediction},  \emph{next friend prediction}, \emph{next evolution prediction} for the next item to be evolved and  \emph{next purchase prediction} for the next item to be bought.

The datasets used in our offline test are samples from the three games. We do not sample users whose behavior sequences are missing. For example, users do not interact with any friend will not be sampled. We show the statistics of datasets in Table~\ref{tab:data}.
Note that, in \lego, players acquire maps and in the rest two games players acquire items. In each dataset, the number of users who are friended is calculated based on users (1) whose behavior sequences are not sampled, and (2) who appear in the friend sequence of a sampled user sequence.

\begin{table}[t]
\caption{Dataset statistics.}\label{tab:data}
\centering
\scalebox{0.9}{
\begin{tabular}{lccccc}
\toprule
Dataset & \#sequence & \#item/map & \#user & \#friend & \#type \\
\midrule
\lego    & 1,884       & 2,446   & 20,319  & 18,435 & 7       \\
T-game  & 60,811      & 62      & 75,518  & 14,707 & 12      \\
Y-game  & 14,571      & 39      & 25,377  & 10,806 & 7      \\
\bottomrule
\end{tabular}
}
\end{table}

\subsection{Evaluation Metrics in Offline Test}
For the offline tests in \mysec{\ref{sec:rq1}} and \mysec{\ref{sec:rq2}}, we evaluate experimental results based on three evaluation metrics in the top-$n$ recommendations, namely Hit Ratio (HR@$n$), Mean Reciptrocal Rank (MRR@$n$), and Normalized Discounted Cumulative Gain (NDCG@$n$). They are commonly adopted in evaluating recommendation performance~\cite{TanXL16,HidasiKBT16,WangHCHLL19,LiuZMZ18,WuT0WXT19}.

\subsection{Contributions of Multi-tasking (\textbf{RQ1})}
\label{sec:rq1}

To understand how each task contribute to \ours's performance (\textbf{RQ1}), we test three multi-task settings.
We investigate the performance of modeling both main task \emph{next map/item prediction} and one auxiliary tasks (i.e., \emph{next type prediction} or \emph{next friend prediction}), and the performance of modeling \emph{all} three tasks. Note that here we only list three tasks that are common in all three games.

We compare different methods for HR@$n$, MRR@$n$, NDCG@$n$, where $n=1,2,\cdots,10$.
We determine the top three results out of eight outcomes 
based on each evaluation metric in each dataset.
Then we score each multi-task setting by calculating the percentage of top-$3$ results: $s_i = \frac{n_i}{N}$, 
where $n_i$ is the number of times that results generated by the $i$-th multi-task setting are the best three results  based on any evaluation metric in any dataset. $N$ is the overall number of top-3 results.

\begin{figure}[t]
\includegraphics[width=0.9\columnwidth]{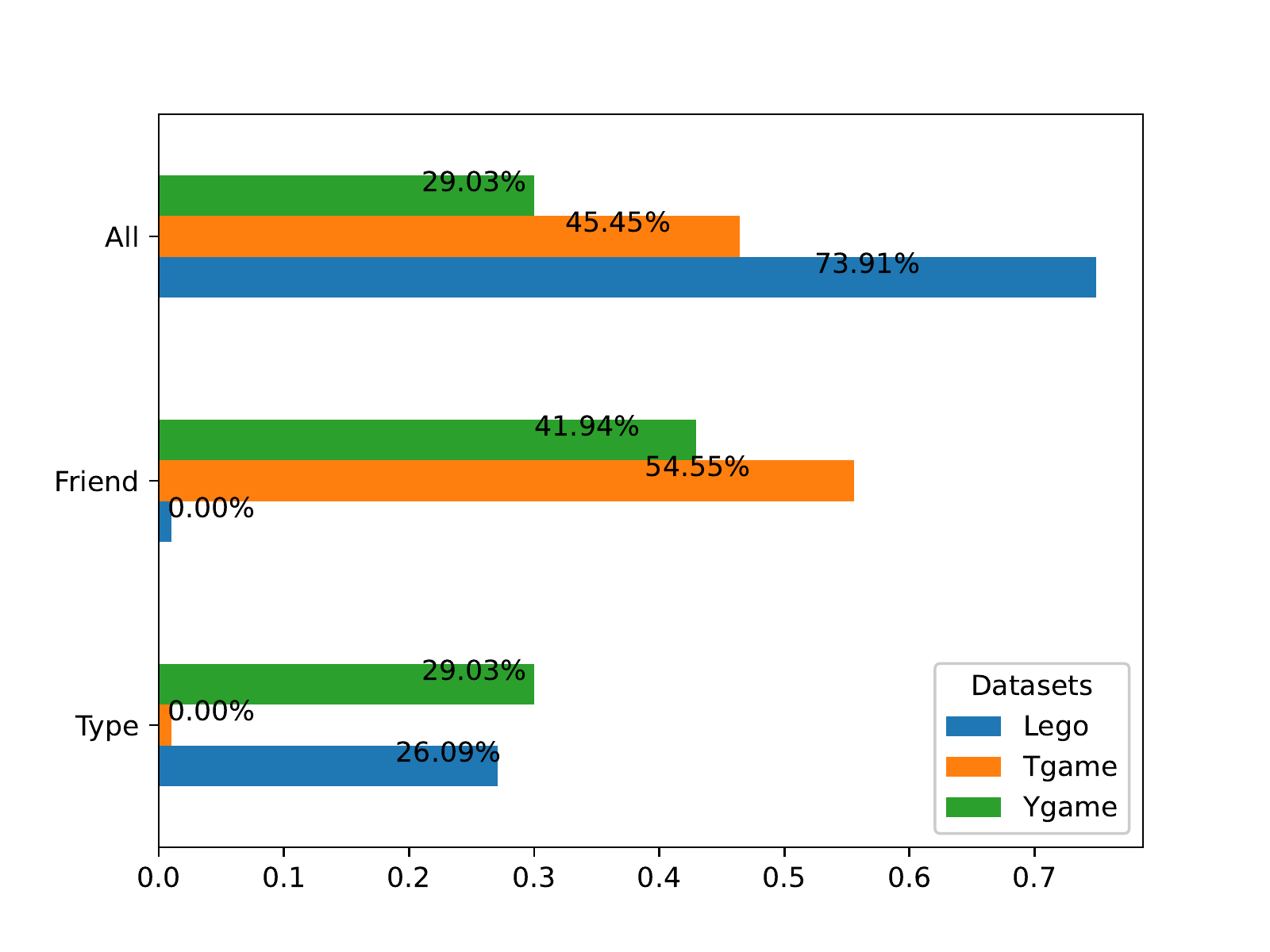}
\caption{Percentage of top-$3$ results generated by different multi-tasking schemes.}\label{fig:variants}
\end{figure}

As shown in \myfig{\ref{fig:variants}}, incorporating all auxiliary tasks generates the best results.
It contributes to $29.03\%$ of the top-3 results in Ygame, $45.45\%$ of the top-$3$ results in Tgame and $73.91\%$ of the top-$3$ results in \lego.
On the contrary, combining only one auxiliary task produces less stable results.
For example, combining the main task with only next friend prediction does not perform well on \lego. Its results are never among the top three best results in terms of any evaluation metric.
Similarly, combining the main task with only next type prediction performs poorly on Tgame, with zero top three results in terms of any evaluation metric.

\begin{table*}[!t]
\centering
\caption{Offline test on \lego dataset with best results in bold.}
\resizebox{0.98\textwidth}{!}{
\begin{tabular}{|l|rrrr|rrr|rrr|}
\hline
Method           & \multicolumn{1}{l}{HR@1} & \multicolumn{1}{l}{HR@2} & \multicolumn{1}{l}{HR@5} & \multicolumn{1}{l|}{HR@10}               & \multicolumn{1}{l}{MRR@2} & \multicolumn{1}{l}{MRR@5} & \multicolumn{1}{l|}{MRR@10}              & \multicolumn{1}{l}{NDCG@2} & \multicolumn{1}{l}{NDCG@5} & \multicolumn{1}{l|}{NDCG@10} \\
           \hline
GRU        & 0.0195                   & 0.0357                   & 0.0737                   & 0.1356                      & 0.0276                    & 0.0378                    & 0.0460                                 & 0.0357                     & 0.0558                     & 0.0770                      \\
MARank     & 0.0151                   & 0.0329                   & 0.0742                   & 0.1311                                  & 0.0240                    & 0.0350                    & 0.0421                                  & 0.0329                     & 0.0547                     & 0.0737                      \\
STAMP      & 0.0184                   & 0.0312                   & 0.0725                   & 0.1278                                  & 0.0248                    & 0.0364                    & 0.0434                                  & 0.0312                     & 0.0539                     & 0.0725                      \\
 SR-GNN        & 0.0234                   & 0.0379                   & 0.0759                   & 0.1088                                  & 0.0307                    & 0.0402                    & 0.0446                                  & 0.0379                     & 0.0572                     & 0.0684                      \\
GAT & 0.0156 & 0.0335 & 0.0670 & 0.1328 & 0.0246 & 0.0334 & 0.0422 & 0.0335 & 0.0511 & 0.0737\\
\hline
\ours   & \textbf{0.0301}          & \textbf{0.0513}          & \textbf{0.0882}          & \textbf{0.1423} & \textbf{0.0407}           & \textbf{0.0505}           & \textbf{0.0575} & \textbf{0.0513}            & \textbf{0.0708}            & \textbf{0.0891}            \\
\hline
\end{tabular}}\label{tab:loffline}
\end{table*}

\begin{table*}[!t]
\caption{Offline test on Tgame dataset with best results in bold.}
\centering
\resizebox{0.98\textwidth}{!}{
\begin{tabular}{|l|rrrr|rrr|rrr|}
\hline
           & \multicolumn{1}{l}{HR@1} & \multicolumn{1}{l}{HR@2} & \multicolumn{1}{l}{HR@5} & \multicolumn{1}{l|}{HR@10}& \multicolumn{1}{l}{MRR@2} & \multicolumn{1}{l}{MRR@5} & \multicolumn{1}{l|}{MRR@10} & \multicolumn{1}{l}{NDCG@2} & \multicolumn{1}{l}{NDCG@5} & \multicolumn{1}{l|}{NDCG@10} \\
           \hline
GRU        & 0.5317                   & 0.6827                   & 0.8544                   & 0.9445                     & 0.6072                    & 0.6546                    & 0.6670                      & 0.6827                     & 0.7760                     & 0.8075                       \\
MARank     & 0.5012                   & 0.6737                   & 0.8532                   & 0.9417                     & 0.5874                    & 0.6369                    & 0.6491                      & 0.6737                     & 0.7710                     & 0.8020                       \\
STAMP      & 0.4798                   & 0.6531                   & 0.8378                   & 0.9337                     & 0.5664                    & 0.6175                    & 0.6307                      & 0.6531                     & 0.7535                     & 0.7870                       \\
SR-GNN        & 0.5327                   & 0.6901                   & 0.8601                   & \textbf{0.9461}                     & 0.6114                    & 0.6587                    & 0.6705                      & 0.6901                     & 0.7830                     & 0.8130                       \\
GAT      & 0.4029          & 0.6094 & 0.8238 & 0.9309 & 0.5061 & 0.5663 & 0.5810 & 0.6094 & 0.7273 & 0.7648 \\
\hline
\ours   & \textbf{0.5328}          & \textbf{0.6910}          & \textbf{0.8605}          & \textbf{0.9461}            & \textbf{0.6115}           & \textbf{0.6589}           & \textbf{0.6706}            & \textbf{0.6910}            & \textbf{0.7834}            & \textbf{0.8133}              \\
\hline
\end{tabular}}\label{tab:toffline}
\end{table*}

\begin{table*}[!t]
\caption{Offline test on Ygame dataset with best results in bold.}
\centering
\resizebox{0.98\textwidth}{!}{
\begin{tabular}{|l|rrrr|rrr|rrr|}
\hline
           & \multicolumn{1}{l}{HR@1} & \multicolumn{1}{l}{HR@2} & \multicolumn{1}{l}{HR@5} & \multicolumn{1}{l|}{HR@10}& \multicolumn{1}{l}{MRR@2} & \multicolumn{1}{l}{MRR@5} & \multicolumn{1}{l|}{MRR@10} & \multicolumn{1}{l}{NDCG@2} & \multicolumn{1}{l}{NDCG@5} & \multicolumn{1}{l|}{NDCG@10} \\
           \hline
GRU        & 0.5556                   & 0.7266                   & 0.9202                   & 0.9882                     & 0.6411                    & 0.6943                    & 0.7037                      & 0.7266                     & 0.8313                     & 0.8552                       \\
MARank     & 0.5545                   & 0.7230                   & 0.9182                   & 0.9863                    & 0.6388                    & 0.6926                    & 0.7021                      & 0.7230                     & 0.8290                     & 0.8528                       \\
STAMP      & 0.5200                   & 0.6867                   & 0.8931                   & 0.9833                    & 0.6034                    & 0.6586                    & 0.6714                      & 0.6867                     & 0.7962                     & 0.8281                       \\
SR-GNN        & 0.5532                   & 0.7305                   & 0.9286                   & \textbf{0.9904}                     & 0.6418                    & 0.6966                    & 0.7053                      & 0.7305                     & 0.8383                     & 0.8602                       \\
GAT & 0.4955 & 0.7031 & 0.9074 & 0.9822 & 0.5993 & 0.6575 & 0.6678 & 0.7031 & 0.8167 & 0.8430\\
\hline
\ours   & \textbf{0.5562}          & \textbf{0.7380}          & \textbf{0.9291}          & 0.9895            & \textbf{0.6461}           & \textbf{0.6987}           & \textbf{0.7077}             & \textbf{0.7380}            & \textbf{0.8420}            & \textbf{0.8634}              \\
\hline
\end{tabular}}\label{tab:yoffline}
\end{table*}

\begin{table*}[!t]
    \caption{Online test on \lego}
    \centering
    \resizebox{0.73\textwidth}{!}{
    \begin{tabular}{lcccccc}
    \toprule
    Method  & NUV & CTR&ER &DR  & CR \\\midrule
    \ours  &$1.00$ &\textbf{$100.00\%\pm3.09\%$} &\textbf{$100.00\%\pm2.69\%$} &\textbf{$100.00\%\pm2.28\%$} &\textbf{$100.00\%\pm2.00\%$}\\
    \midrule  
    IMF & $0.12$ & $67.32\%\pm7.32\%$ &$70.29\%\pm6.53\%$ &$70.44\%\pm5.61\%$ &$71.41\%\pm5.07\%$\\  
    SNS+IMF & $0.12 $ &$58.05\%\pm6.83\%$ &$66.58\%\pm6.40\%$ &$70.06\%\pm5.61\%$ &$67.79\%\pm4.92\%$\\
    RNN & $1.31$  &$92.20\%\pm2.60\%$ &$96.16\%\pm2.30\%$ &$96.48\%\pm2.00\%$ &$95.08\%\pm1.69\%$ \\
    HERec &$1.10 $  &$73.01\%\pm2.44\%$ &$75.29\%\pm2.18\%$ &$72.53\%\pm1.90\%$ &$72.71\%\pm1.69\%$\\ 
    oKNN  & $5.46 $ &$77.56\%\pm1.14\%$ &$75.16\%\pm1.02\%$ &$75.38\%\pm0.86\%$ &$78.17\%\pm0.77\%$\\   
    Random  &$0.60$ &$65.53\%\pm3.25\%$ &$44.56\%\pm2.30\%$ &$53.61\%\pm2.19\%$ &$58.42\%\pm2.08\%$\\ 
    \bottomrule
    \end{tabular}
    }
    \label{tab:lonline}  
\end{table*}

\begin{table}[!t] 
    \caption{Online test on Tgame}
    \centering
    \scalebox{1.02}
    {
        \begin{tabular}{lccc}
        \toprule
        Method   & NUV & ARPU & CR($\%$) \\
        \midrule
        \ours & $1.00$                 & \textbf{$1.00\pm 0.02$}           & \textbf{$100.00\pm 1.05$}             \\
        \midrule
        eALS     & $0.95$                 & $0.84\pm 0.02$           & $93.98 \pm 1.10$             \\
        RF       & $0.94$                 & $0.92 \pm 0.02$          & $91.31 \pm 1.12$             \\
        IMF      & $1.30$                 & $0.79 \pm 0.02$          & $82.96 \pm 0.97$             \\
        Random   & $0.93$                 & $0.21 \pm 0.01$          & $32.20\pm 0.94$             \\
        \bottomrule
        \end{tabular}
    }
    \label{tab:tonline}
\end{table}

\begin{table}[!t]
\caption{Online test on Ygame}
\centering
    \scalebox{1.15}
    {
        \begin{tabular}{lccc}
        \toprule
        Method   & NUV & ARPU & CR($\%$) \\
        \midrule
        \ours & $1.00$                 & \textbf{$1.00\pm 0.46$}           & \textbf{$100.00\pm 20.62$}             \\
        \midrule
        IMF     & $5.19$                 & $0.60\pm 0.13$           & $68.33 \pm 7.50$             \\
        POP       & $3.05$                 & $0.19 \pm 0.07$          & $41.60 \pm 7.65$             \\
        \bottomrule
        \end{tabular}
    }
\label{tab:yonline}
\vspace{10pt}
\end{table}

\subsection{Offline Comparative Performance (\textbf{RQ2})}
\label{sec:rq2}
To answer \textbf{RQ2}, we compare \ours with the following methods as baselines in the offline test:
\begin{itemize}
\item \ul{GRU}~\cite{TanXL16} is a state-of-the-art sequential recommendation model which applies Recurrent Neural Networks (RNN) with ranking-based loss functions.  It improves~\cite{HidasiKBT16} by data augmentation, and a method to account for shifts in the input data distribution.
\item \ul{MARank}~\cite{WangHCHLL19} is recently proposed to unify both individual- and union-level item interaction into preference inference model from multiple views. It utilizes attention mechanism and residual neural network.
\item \ul{STAMP}~\cite{LiuZMZ18} presents a novel short-term attention/memory priority model to capture users' general interests from the long-term memory of a session context, whilst takes into account users' current interests from the short-term memory of the last action.
\item \ul{SR-GNN}~\cite{WuT0WXT19} models sequences as graph data and applies Gated Graph Neural Network~\cite{LiTBZ15} to extract complex transitions of items. An attention network is utilized to represent each sequence as a composition of global preference and current preference from the last action.
\item \ul{GAT}~\cite{Qiu2019Rethinking} collaboratively considers the sequence order and the latent order of user preference by constructing a session graph and utilizing a weighted attention graph layer.
\end{itemize}
All methods, including competitors and \ours use an embedding size of 128, a learning rate of 0.0001 and a batch size of 128. For other special hyper-parameters of competitors, we use the recommended default settings given in their papers or open-source implementations.

The results of HR@$n$, MRR@$n$, and NDCG@$n$ with $n=1,2,5,10$ on the three datasets are shown in \mytabs{\ref{tab:loffline},~\ref{tab:toffline} and~\ref{tab:yoffline}}. We omit MRR@1 and NDCG@1 as these two are identical with HR@1. 
For \ours, the best performance multi-task setting is reported. 
We can observe that \ours performs consistently best in terms of all evaluation metrics on all datasets.
This shows that \ours obtains very promising accuracy in the top-$n$ sequential recommendations.

\subsection{Online Comparative Performance (\textbf{RQ3})}
To validate that \ours has indeed promoted recommendation performance (\textbf{RQ3}), we
conduct an online evaluation through a A/B test. We randomly select a group of
exposed users for each recommendation approach (e.g., the competitors or
\ours). Then we deliver top items with highest prediction score by each
recommendation approach to the exposed users on the main task. 

\vspace{5pt}
\noindent\textbf{Competitors.} Note that the competitors used
in the online test are currently deployed in the online games. Due to various
reasons such as security and efficiency, the recommender of each game might
have differing models which are also different compared to those used in our
offline test.
\begin{enumerate}
\item \textbf{Competitors on \lego}:
\vspace{5pt}
\begin{itemize}
\item \ul{IMF}~\cite{Hu2008Collaborative} is a latent factor model treating the data as positive and negative instances with vastly varying confidence levels. 
\item \ul{SNS+IMF} improves IMF by leveraging the social network information. The confidence of each instance in the social network is computed by the efficient Personalized PageRank (PPR) algorithm on large graphs using the distributed computing framework~\cite{Lin19WWW}. 
\item \ul{HERec}~\cite{ShiHZY19} is a state-of-the-art recommendation model based on heterogeneous information network. 
\item \ul{RNN}~\cite{HidasiKBT16} deploys RNNs to predict next map. 
\item \ul{oKNN}~\cite{abs-0712-4273} is the online KNN algorithm, where each recommendation is made based on the majority voting from the user's cluster. 
\item \ul{Random} is to randomly predict the next map.
\end{itemize}
\vspace{5pt}
\item \textbf{Competitors on Tgame.} 
\vspace{5pt}
\begin{itemize}
\item \ul{eALS}~\cite{He2016Fast} adopts an efficient element-wise Alternating Least Squares (eALS) learning technique in factorization machine~\cite{KorenBV09} which weighs the missing data based on item popularity.
\item \ul{RF}~\cite{Liaw2002Classification} is the classic random forest prediction model.
\item \ul{IMF}~\cite{Hu2008Collaborative} is the latent factor model which learns from weighted, positive and negative instances.
\item \ul{Random} is to randomly predict the next item.
\end{itemize}
\vspace{5pt}
\item \textbf{Competitors on Ygame} include \ul{IMF} as used in the other two games and \ul{POP} which is to give the most popular item as the prediction for the next item.
\end{enumerate}

\vspace{5pt}
\noindent\textbf{Evaluation Metrics.} The evaluation metrics are suggested by the operation team in Tencent game, which are designed for each game in the purpose of better operation. The following evaluation metrics are based on $UV$, which is the number of users exposed in each group:
\begin{enumerate}
\item \textbf{Evaluation Metrics for \lego}:
\vspace{5pt}
\begin{itemize}
\item \ul{Click Through Rate (CTR)} is calculated as $CTR=C/UV$, where $C$ is the total clicks received from the exposed users.
\item \ul{Download Rate (DR)} is calculated as $DR=D/UV$, where $D$ is the number of exposed users who download the recommended map.
\item \ul{Entrance Rate (ER)} is calculated as $ER=E/UV$, where $E$ is the number of exposed users who enter the recommended map in \lego.
\item \ul{Conversion Rate (CR)} is calculated as $CR=PU/UV$, where $PU$ is the number of purchasing users.
\end{itemize}
\vspace{5pt}
\item \textbf{Evaluation metrics on Tgame and Ygame} include \ul{Conversion Rate (CR)} as used in \lego and \ul{Average Revenue Per User (ARPU)} (i.e., $ARPU=R/UV$, where $R$ is the total revenue and $UV$ is the number of users exposed).
\end{enumerate}

As an effort to quantifying the size of each experiment group without revealing the exact number of users exposed, we also give the \ul{Normalized User View (NUV)}. NUV is the number of users exposed in each experiment group divided by the number of users exposed in the group for \ours. Thus the NUV of \ours always equals one.

We report the results with the confidence level $rho\geq 0.95$ in
\mytabs{\ref{tab:lonline},~\ref{tab:tonline} and~\ref{tab:yonline}}. We can observe that \ours
significantly outperforms previously deployed competitors, in terms of all
evaluation metrics that are designed by the operation team. This clearly
suggests the great potential of \ours in online gaming.


\section{Related Work} 
\label{sec:rel}

In this section, we will elaborate on
the relevant works which can be concluded into three main paradigms:
\emph{General Recommender Systems}, \emph{Sequential Recommender Systems}
and \emph{Multi-task Recommender Systems}.

\subsection{General Recommender Systems} 

Recommender Systems (RS) have become
an essential tool for solving information overload problem~\cite{2015rsh}. RS
not only assists users in searching for desirable targets but also helps
e-commerce platforms promote their products and boost sales~\cite{Aggarwal16}.
Traditional RS do not consider sequential behaviors and they
typically rely on collaborative filtering methods (CF), especially matrix
factorization (MF)~\cite{KorenBV09}, to utilize historical user-item
interactions for recommendation. MF factorizes the user-item interaction
matrix into two low-dimensional latent matrices while preserving the inherent
information from original user-item interactions. MF has proven its ability of
modeling user preferences and item properties in Netflix Prize
Challenge~\cite{BellK07}. Due to its effectiveness when handling large-scale
data~\cite{LiCYM17}, MF has been successfully deployed in the industry (e.g.,
Facebook, Amazon and Netflix~\cite{KorenBV09,Aggarwal16}). One of the most
challenging issues in traditional recommender systems is the cold-start
problem, where historical data is not available for new users or
items~\cite{Aggarwal16}. In order to alleviate the cold-start problem, many
works have incorporated additional context information, which is also called
auxiliary data or side information (e.g., social network~\cite{LiWM14,LiWTM15},
review text~\cite{WangNL2019,Garcia-DuranGON20}, image~\cite{WangNL18}, structural data~\cite{LiLQMTC19,DingLHM17}, and
location~\cite{LuLMC17}), into recommendation models. However, traditional
recommender systems only consider user-item interactions. It is difficult to
use general recommenders for sequential recommendation task directly, since
user sequential behaviors should also be modeled~\cite{LiRCRLM17}.

\subsection{Sequential Recommender Systems} \label{sec:sbrs} 

The research of sequential recommender systems (SRS) has emerged recently, as many
real-world applications have sequence based
traits~\cite{LudewigJ18,abs-1902-04864}.

The pioneering works for sequential recommender systems (SRS) utilize Markov
Chain (MC) which views the three-way interactions in SRS as two components: one
is the interaction between the user and the next item and the other is the
sequential history between previous items and the next
item~\cite{ZimdarsCM01,ShaniBH02, ShaniHB05,ChenMTJ12,TavakolB14}. The former
is well studied in Matrix Factorization (MF)~\cite{KorenBV09} and the later
can be sequentially modeled by MC. Therefore, MC based methods adopt the ideas
from both MF and MC. The drawback of MC based model is that the state space
quickly becomes unmanageable when trying to include all possible sequences of
user selections~\cite{HidasiKBT16}. Another line of works extends MC based SRS using the idea of ``translation'': the next item is viewed as the translation from previous item via the user~\cite{HeKM17,LiLMR20,Garcia-DuranGON20}. The major advantage of these works is that they are much faster than other methods. There are other works using non-machine learning methods. For example, Migliorini et al.~\cite{MiglioriniQCB19} deals with the production of sequences of recommendations for dynamic groups by considering the role of the context.

Hidasi et al.~\cite{HidasiKBT16} firstly introduce Recurrent Neural Network
(RNN) into sequential recommendation problem and proposed GRU4Rec which
utilizes RNNs with a Gated Recurrent Unit (GRU) for SRS. Later, they extended
GRU4Rec to exploit additional features (e.g., picture and text) by using
parallel RNN architectures~\cite{HidasiQKT16}. Due to the large performance
gain of GRU4Rec over traditional methods, RNN based methods (including RNN,
GRU and LSTM) have become prevalent in recent studies of
SRS~\cite{TanXL16,Twardowski16,SmirnovaV17,ChatzisCA17,LiRCRLM17,LoyolaLH17,JannachL17,WangCZLL18,WuY17,HidasiK18,abs-1812-02646}.
In addition to RNN, a few researchers explored the possibility of using other
neural networks for SRS. Tuan and Phuong~\cite{TuanP17} harnessed 3D
Convolutional Neural Network (CNN) and side information to enhance the
accuracy of sequential recommendation. Wu et al.~\cite{WuT0WXT19}
investigated how to improve SRS with Graph Neural Network (GNN) and each
sequence is then represented as the composition of the global preference and
the current interest of that sequence using an attention network. Following Wu et al.~\cite{WuT0WXT19}, a few recent works explore the potential of GNN in SRS~\cite{Qiu2019Rethinking,XuZLSXZFZ19,QiuYHC20}.

There are also some
SRS considering the availability of user identity in each sequence. Due to the
fact that the past and current user sequences can be simply concatenated to
obtain longer sequence for the same user.
Epure et al.~\cite{EpureKIDSA17} introduced the
concept of medium-term behavior in addition to the existing short-term and
long-term behaviors in personalized SRS for news and combined them together
to enhance recommendation on the top of MC.  Quadrana et
al.~\cite{QuadranaKHC17} argued that concatenating user sequences when user
identity is available will not yield the best result and proposed a
Hierarchical RNN model with cross-session information transfer. Inspired by
the success of language modeling, Hu et al.~\cite{HuCWXCG17} modeled the
information of the user and items in a sequence as the context and used
probabilistic classifier to identify the item to be recommended next.  Liu et
al.~\cite{LiuZMZ18} noticed the user interests drift in a long user sequence is
not well modeled by conventional models. Therefore, they designed a short-term
attention/memory priority model as a remedy. Song et al.~\cite{Song0WCZT19}
modeled social influence in sequential social recommender with RNN and a
graph-attention neural network.

As explained in \mysec{\ref{sec:intro}}, conventional SRS are single-sequence
and single-task and hence they are not suitable for the multi-sequence and
multi-task recommendation task in Tencent Games platform.

\subsection{Multi-task Recommender Systems} Multi-task learning has been
successfully deployed in several applications~\cite{ZhangY17aa}. Multi-task
learning aim is to leverage useful information contained in multiple related
tasks to help improve the generalization performance of all the tasks. 

As far as we know, multi-task learning has not been introduced to SRS before,
but there are some efforts in using multi-task learning to improve other types
of RS. Wang et al.~\cite{WangHZL13} proposed OMTCF which models each user in
online CF as an individual task. OMTCF not only update the weight vectors of
the user (task) related to the current observed data, but also the weight
vectors of some other users (tasks) according to a user interaction matrix.
With the similar idea, Wang et al.~\cite{WangHZL13} introduced a multi-task
learning framework which learns multiple rating prediction models
simultaneously (one for the active user and one for each of the related
users). Chen et al.~\cite{abs-1901-06125} designed a multi-task framework for
music playlist recommendation, which can deal with three recommendation tasks
(i.e., cold playlist recommendation, cold user recommendation and cold song
recommendation) in parallel. Ni et al.~\cite{NiOLLOZS18} considered optimizing
multiple search and recommendation tasks in e-commerce platform and learned
universal user representations across multiple tasks for more effective
personalization. Multi-task learning is also used in knowledge graph enhanced
RS. For instance,  knowledge graph completion task can be utilized to assist
recommendation task~\cite{abs-1901-08907,abs-1902-06236}. Moreover, the
induction of explainable rules from knowledge graphs can be integrated with
recommendation task. Ma et al.~\cite{abs-1903-03714} and the two tasks
complement each other in a multi-task framework.


\section{Conclusion}
\label{sec:con} 

In this paper, we study a new research problem which naturally arises from the
recommendation scenario of online games. Using the ideas of GNN and multi-task
learning, we propose a new method \ours to fully utilize the complex
information in the heterogeneous sequential recommendation scenario of online
games. Online evaluations in three games of Tencent Games platform
illustrate the effectiveness of \ours. In the future, we plan to enhance the interpretability of \ours so that we can better understand the recommendation results.

\begin{acks} 
Chen Lin is supported by the Natural Science Foundation of
China (No. 61972328) and Joint Innovation Research Program of Fujian Province China (No. 2020R0130).
Hui Li is supported by the Natural Science Foundation of China (No. 62002303) and Natural Science
Foundation of Fujian Province China (No. 2020J05001).
\end{acks}

\bibliographystyle{ACM-Reference-Format}
\bibliography{ref.bib}

\end{document}